\documentclass[a4paper,11pt,onecolumn]{article}
\usepackage[english]{babel}
\usepackage[utf8]{inputenc}
\usepackage{bibgerm}
\usepackage{graphicx}
\usepackage{hyperref}
\usepackage{authblk}
\usepackage[top=2cm, bottom=3cm, left=2cm, right=2cm]{geometry}

\title{Entrainment profiles: Comparison by gender, role, and feature
  set}
\author{{\em Uwe D. Reichel $^a$, \v{S}tefan Be\v{n}u\v{s} $^{b,c}$, Katalin M\'ady $^d$}}
\affil{{\em $^a$University of Munich, Germany}\\{\em $^b$Constantine the Philosopher University, Nitra, Slovakia}\\{\em $^c$Institute of Informatics, Slovak Academy of Sciences, Bratislava, Slovakia}\\{\em $^d$Hungarian Academy of Sciences, Budapest, Hungary}}
\date{}

\begin{document}

\maketitle
  
\hrule

\begin{abstract}  
  We examine prosodic entrainment in cooperative game dialogs for new
  feature sets describing register, pitch accent shape, and rhythmic
  aspects of utterances. For these as well as for established features
  we present entrainment profiles to detect within- and across-dialog
  entrainment by the speakers' gender and role in the game.  It turned
  out, that feature sets undergo entrainment in different quantitative
  and qualitative ways, which can partly be attributed to their
  different functions. Furthermore, interactions between speaker
  gender and role (describer vs. follower) suggest gender-dependent
  strategies in cooperative solution-oriented interactions: female
  describers entrain most, male describers least. Our data suggests a
  slight advantage of the latter strategy on task success. \\

  \vspace*{-0.5cm}

  \begin{flushleft}
  \end{flushleft}

\end{abstract}

\hrule

\vspace*{1cm}

\begin{flushleft}
\begin{small}
{\bf Accepted Manuscript} for {\em Speech Communication} (Elsevier), 25 April 2018
\\ 
{\bf Reference:} U.D. Reichel, \v{S}. Be\v{n}u\v{s}, and
K. M\'ady. Entrainment profiles: Comparison by gender, role, and
feature set. {\em Speech Communication}, 100:46--57, 2018.\\
{\bf DOI:} \url{https://doi.org/10.1016/j.specom.2018.04.009}
\end{small}
\end{flushleft}

\newpage

\section{Introduction}
\label{sec:intro}

In spoken conversations, multiple aspects of interlocutors' utterances
and their speaking behavior tend to become more similar to each
other. This phenomenon is called entrainment in the computer science
literature and is also commonly referred to as alignment,
accommodation, audience design, mimicry, priming, or other in
psychology, sociology and other disciplines. There are several well
established and relatively non-controversial aspects of
entrainment. First, entrainment affects not only speech but also other
modalities such as gaze, facial expression, mannerisms, or posture
\cite{Chartrand1999}. In this paper, we concentrate only on
entrainment in the speech modality as entrainment in most studies was
observed in spoken interactions (and possibly even without visual
contact between interlocutors \cite{Shockley2003}), which points to
speech as playing an important and natural role for entrainment also
in other modalities.

Second, entrainment affects both linguistic and para-linguistics
domains of speaking. On the linguistic level entrainment affects
amongst others the choice of words \cite{Nenkova2008, Danescu2012,
  BrennanEP1996} or syntactic constructions \cite{Cleland2003,
  Gries2005, BraniganC2007}. While the text/transcript discrete data
are predominantly used for analyzing the linguistic aspects, the
continuous acoustic-prosodic features extracted directly from the
speech signal have been commonly used to explore entrainment in the
para-linguistic domain (speech rate, intensity, pitch, voice quality
\cite{Gregory1996, Gregory1997, Levitan2011, Levitan2012,
  Babel2012}). A notable exception is the study analyzing entrainment
in terms of linguistically meaningful aspects of intonational contours
via discrete ToBI labeling \cite{GravanoSLT2014}.

Third, speech entrainment tends to correlate with positive perception
of the interlocutor and/or interactions in which entrainment took
place. Entrainment has been shown to increase the success of
conversation in terms of low inter-turn latencies and a reduced number
of interruptions \cite{Levitan2012, Nenkova2008} as well as with
objective task success measures \cite{Reitter2014}, and people are
generally perceived as more socially attractive and likable, more
competent and intimate if they entrain to their interlocutors (reviews
in \cite{Hirschberg2011} and \cite{BenusCC2014}). More recently,
entrainment was also found to play an important role in the perception
of social attractiveness and likability \cite{Schweitzer2017,
  Michalsky2017b} This extends also to some aspects of human-machine
spoken interactions in which bi-directional entrainment between humans
and machines improved the effectiveness and user's experience of the
interactions (review in \cite{BenusCC2014}) and several approaches are
proposed for endowing synthesizers with speech entrainment
capabilities \cite{Levitan2016, Lubold2015, Raveh2017}.

However, recent research also suggests that the link between speech
entrainment and aspects characterizing spoken interaction is more
complex. First, as also pointed out by an anonymous reviewer, a causal
link between entrainment and task success has not been clearly
established and the observed positive correlations may stem from a
stronger social relationship reflected by greater collaboration,
engagement, and/or entrainment. Moreover, several studies also
suggest that both entrainment and disentrainment co-occur integrally
in conversations and that positive aspects perceived in the
interactions may be linked to their combination \cite{Perez2016,
  LoozeSC2014, Healey2014}. This complexity is further corroborated by
studies showing that convergence and synchrony in pitch features have
complex and complementary relationships with the speaker’s impression
of their interlocutor's visual attractiveness and likability
\cite{Michalsky2017a, Michalsky2017b}.

In addition, there are some other aspects of entrainment that are
still not well understood. The first general issue of contention in
cognitive science and psychology is the degree of control a speaker
has over entrainment to her interlocutor. Despite differences, two
influential approaches to entrainment (\cite{Pickering2004,
  Pickering2013} and \cite{Chartrand1999}) suggest that entrainment is
in general an automatic priming-type mechanism rooted in the
perception-production link in which the activation of the linguistic
representations or other behavior from the interlocutor increases the
likelihood of producing such representations/behavior by the speaker.
On the other hand, the Communication Accommodation Theory (CAT)
\cite{Giles1991} maintains that speakers use entrainment or
dis-entrainment in order to attenuate (or accentuate) social
differences and thus actively negotiate social distance in spoken
interactions. Several studies propose a hybrid approach in which the
link between processes of perception and production is not automatic,
but can be mediated by pragmatic goals or social factors
\cite{Kraljic2008, Schweitzer2014}.

A more specific issue, that is directly relevant to the first one,
involves the role of gender and power relations of the
interlocutors. Entrainment turns out to be stronger in case of mutual
positive attitude of the interlocutors, than in case of negative
attitude \cite{Lee2010}, which is in line with the predictions of
theoretical models such as the CAT \cite{Giles1991}. The CAT also
predicts a dependence of entrainment on dominance relations. In case
of a misbalanced power of two interlocutors the one with the lower
status (or authority, dominance) will entrain more to the one with the
higher status \cite{Giles2007}. Empirical evidence for this claim has
been found amongst others for talkshow data \cite{Gregory1996}, the
judicial domain \cite{Danescu2012}, or in task-oriented dialogs
\cite{BenusJP2011}, where hierarchies turned out to be well reflected
in the amount of entrainment. Combining this with the male-dominance
hypothesis \cite{Zimmermann1975}, we may hypothesize that female
speakers generally entrain more than males.  In addition to this
sociological reasoning, greater entrainment of females compared to
males might be hypothesized based on the above mentioned link between
entrainment and the perception-production loop: females might be
capable to entrain more, since they are more sensitive to fine
phonetic detail than males \cite{Chartrand1999}.

Support has been found for both the male-dominance hypothesis in terms
of higher frequencies of interruptions and ego first-person singular
pronouns \cite{Zimmermann1975}, and for the higher phonetic
sensitivity of female speakers \cite{Namy2002}.

However, the picture of gender-related entrainment differences is much
less clear than to be expected based on the literature. Some studies
explore only mixed-gender dyads \cite{Michalsky2017a}, others
\cite{Bilous1988, XiaSP2014} revealed complex patterns of
gender-related entrainment in same- and mixed-gender dyads that are
furthermore feature- and language-dependent. Similarly, the interplay
between gender and the conversation role on entrainment is not
clear. \cite{BraniganC2007} for example analyzed data from multi-party
picture-describing task in which the degree of syntactic entrainment
of the participants in a current picture-description was affected by
the speaker's role in the previous description (addressee or
side-participant) but not by the addressee's role. However, the gender
of the participants is not specified in this study and these
conversational roles do not yield straightforwardly to power
differences.

Another specific issue involves the type of features commonly used
in entrainment research. Since the linguistic features require
transcripts and (shallow) parsing or expensive annotation of the data
(e.g. ToBI labeling), studies exploring entrainment based only on the
signal focused on coarse acoustic-prosodic (a/p) features. This makes
sense also for applied research since the upshot of understanding
speech entrainment in human-human spoken interactions is in designing
interactive spoken dialog systems with online entrainment
capabilities so that human-machine spoken dialog systems in the
future are more effective and more positively perceived by humans. The
coarse a/p features are easily extractable from the signal and can be
in turn easily adaptable in speech synthesis for entrainment
purposes. However, the speech signal may also contain automatically
extractable information about higher-level features that are
intermediate between para-linguistic and linguistic and include, for
example, features characterizing the shape of intonational contours in
relevant speech intervals. Analyzing the relevance of such features
for entrainment, and their relationship to the traditional a/p
features will fill the current gap in our understanding of speech
entrainment.

\paragraph{Goals of the current study}
We will address the two specific issues mentioned above by
disentangling the gender and communicative role in analyzing how they
participate on entrainment. That is, we will not predefine male and
female authority, as a special case of 'power', in terms of the
male-dominance hypothesis, but assign it to the speaker's role in a
cooperative game.  Technically, in order to examine entrainment
selectively by speaker role and gender we propose an asymmetric turn
pairing procedure that yields separate entrainment values for each
speaker. We also will address a potential impact of entrainment
behavior by role and gender on task success. Furthermore, we will
extend the prosodic feature pool to be investigated. All pitch
examinations cited above were restricted to rather coarse acoustic
measures such as the mean or maximum value of the fundamental
frequency (f0) \cite{Levitan2011, Levitan2012}, its variance
\cite{Gregory1996} and the distance between raw f0 contours
\cite{Babel2012}. We will add features derived from a parametric
superpositional intonation stylization, that allow for the comparison
of more complex pitch patterns in different prosodic domains. These
contextualized features furthermore allow for a positional examination
of entrainment, that is, whether more entrainment occurs in the
beginning or the end of a turn.

Finally, although we introduce some new a/p features and factors
(role/ gender) in exploring entrainment, we strive to make our results
comparable to the existing literature by basing our quantification of
entrainment on the notions of synchrony and proximity. In this we
follow previous studies \cite{Edlund2009, Levitan2011, LoozeSC2014}
that explored the signal-based continuous features. Another line of
alignment/ accommodation research bases their analyses on discrete
data and are largely dependent on quantifying the ratios or
probabilities of exact repetitions of certain text-based linguistic
structures or lexical items \cite{Fusaroli2012, Xu2015, Jones2014}. We
leave the comparison of the signal-based and text-based
operationalizations of entrainment for future research.

After the presentation of our data and the extracted prosodic features
(sections \ref{sec:data} and \ref{sec:feat}) we will introduce
profiles of several operationalizations of entrainment (section
\ref{sec:prof}). The observations obtained from these profiles will be
tested and discussed in sections \ref{sec:harv} and \ref{sec:disc}.

\section{Data}
\label{sec:data}

\subsection{Corpus}

The Slovak Games Corpus (SK-games) was used;
e.g. \cite{Benus2016}. The corpus was recorded with slight
modifications following the Object games of the Columbia Games Corpus
\cite{Gravano2009, GravanoHirschberg2011}. Briefly, pairs of subjects
were seated in a quiet room opposite each other but without any visual
contact and used the mouse to move images on the screens from their
initial positions to the target positions. One of the subjects saw the
target position on her screen (the Describer) and guided the other
player (the Follower) to place the image into that position. The
players were awarded points based on a pixel-match between the target
position on the Describer's screen and the placement on the Follower's
screen. In each session the subjects placed 14 images and they
regularly switched roles of the Describer and Follower. This design
resulted in natural task-oriented collaborative dialogs.  The
material comprises 9 sessions of approximately 6 hours of dialogs by
11 speakers (5 female, 6 male; 5 mixed gender, 2 female-female, and 2
male-male dialogs).

\subsection{Preprocessing}

\paragraph{Alignment}

The manually derived text transcription within the
semi-\-auto\-mati\-cal\-ly determined inter-pausal units (IPUs,
threshold of 100ms) was automatically aligned to the signal on the
sound and word levels using the SPHINX toolkit adjusted for Slovak
\cite{Darja2011}. This forced alignment occasionally produced short
silent periods within the originally determined IPUs and the entire
alignment was manually corrected by the second author.

\paragraph{F0 and energy}

F0 was extracted by autocorrelation (PRAAT 5.3.16 \cite{Boersma1999},
sample rate 100 Hz). Voiceless utterance parts and f0 outliers were
bridged by linear interpolation. The contour was then smoothed by
Savitzky-Golay filtering \cite{Savitzky1964} using third order
polynomials in 5 sample windows and transformed to semitones relative
to a base value. This base value was set to the f0 median below the
5th percentile of an utterance and serves to normalize f0 with respect
to its overall level.

Energy in terms of root mean squared deviation was calculated with the
same sample rate as f0 in Hamming windows of 50 ms length.

\paragraph{Prosodic structure}

The dialogs were segmented into turns and interpausal units. The
latter we employed as a coarse approximation of prosodic phrases given
that speech pauses are among the most salient phrase boundary cues
\cite{Swerts1994}. By this simplifying assumption we use the terms
``interpausal unit'' and ``prosodic phrase'' interchangeably in the
following.  Automatic syllable nucleus assignment follows the
procedure introduced in \cite{Pfitzinger1996} to a large extent. An
analysis window $w_a$ and a reference window $w_r$ with the same time
midpoint were moved along the band-pass filtered signal in 50ms
steps. Filtering was carried out by a 5th order Butterworth filter
with the cutoff frequencies 200 and 4000Hz. For a syllable nucleus
assignment the energy in the relevant frequency range $r$ is required
to be higher in $w_a$ than in $w_r$ by a factor $v$, and additionally
had to surpass a threshold $x$ relative to the maximum energy
$\textrm{RMS}_{max}$ of the utterance, i.e. $\textrm{RMS}(w_a) >
\textrm{RMS}(w_r) \cdot v \land \textrm{RMS}(w_a) > \textrm{RMS}_{max}
\cdot x$. Based on the tuning results in \cite{reichelESSV2017} the
parameters were set to the following values: $w_a=0.05s$, $w_r=0.11s$,
$v=1.1$, $x=0.1$.

Pitch accents were detected automatically by means of a bootstrapped
nearest centroid classifier as described in detail in
\cite{reichelESSV2017}. Based on pitch accent-related features derived
for each word-initial syllable introduced in section \ref{sec:loc} as
well as by vowel length z-scores, two centroids for accented and
non-accented syllable were bootstrapped based on two simplifying
assumptions: (1) all words longer than a threshold $t_a$ are likely to
be content words that contain a high amount of information and are
thus taken as class 1 (accented) representatives, and (2) all words
shorter than a threshold $t_{na}$ are likely to be function words with
a low amount of lexical information and are thus taken as class 0 (no
accent) representatives. $t_a$ and $t_{na}$ were set to 0.6s and
0.15s, respectively. For words fulfilling criterion (1) the first
syllable (Slovak has fixed word-initial stress) was added to the class
1 cluster. For words fulfilling criterion (2) all syllables were added
to the class 0 cluster.

From this initial clustering feature weights were calculated from the
mean cluster silhouette derived separately for each feature. The
weights thus reflect how well a feature separates the seed clusters.

After this cluster initialization the remaining word-initial syllables
are assigned to the classes 0 or 1 in a single pass the following way:
for each feature vector $i$ its weighted Euclidean distances $d_{i,0}$
and $d_{i,1}$ to the class 0 and class 1 centroids are calculated, and
the quotient of both distances $q_i=\frac{d_{i,0}}{d_{i,1}}$ is
recorded. All items with a $q_i$ above a defined percentile $p$ are
assigned to class 1, and the items below to class 0. By choosing a
percentile threshold well above 50 the skewed distribution of class 0
and class 1 cases for both boundaries and accents can be tackled,
i.e. more items receive class 0 than class 1. The percentile threshold
$p$ was set as in \cite{reichelESSV2017} to 82.

In \cite{reichelESSV2017} this procedure yielded F1 scores up to 0.63
on spontaneous speech data, which clearly indicates moderate precision
and recall values for pitch accent detection. However, the choice of
the feature sets for accent detection ensures that syllables with
salient pitch and energy movements are identified for further
analyses.

\section{Prosodic features}
\label{sec:feat}

\begin{figure}
  \begin{center}
    \includegraphics[width=12cm]{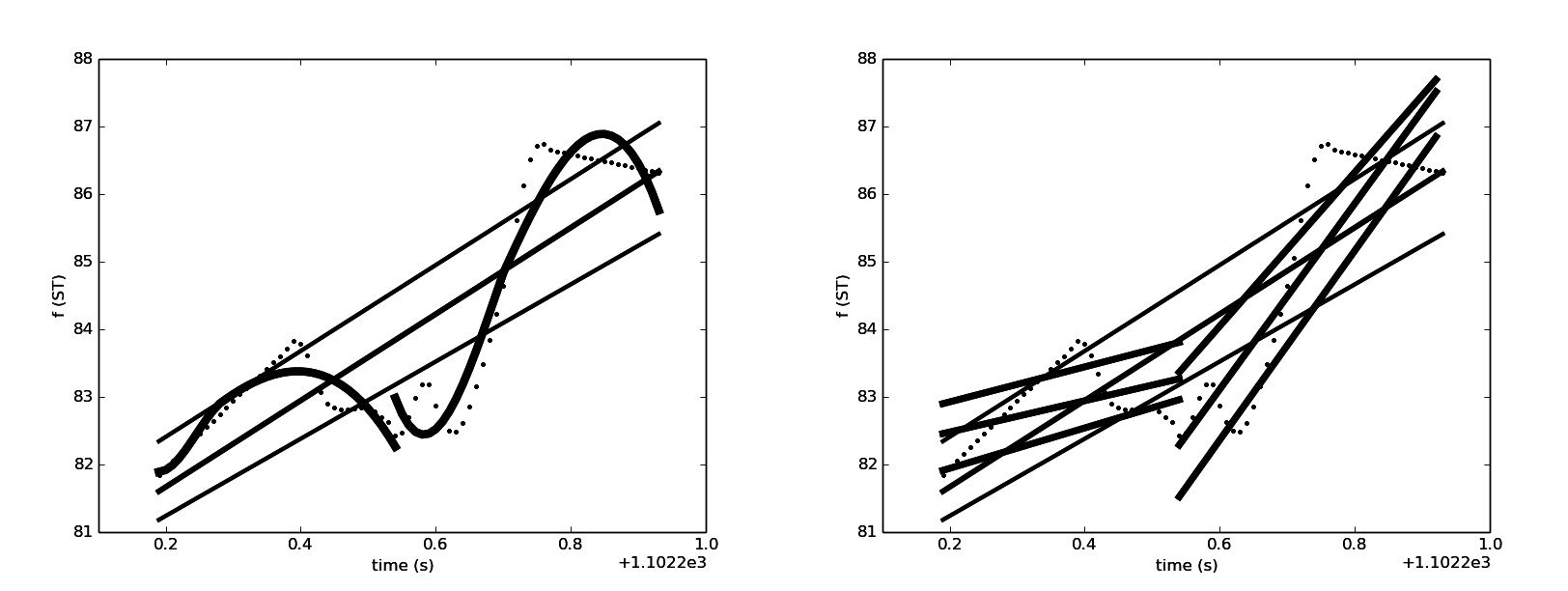}
    \caption{Superpositional f0 stylization within the CoPaSul
      framework. On the interpausal unit (IPU) level a base, mid- and
      topline (solid) are fitted to the f0 contour (dotted) for
      register stylization. Level is represented by the midline, range
      by a regression line fitted to the pointwise distance between
      base and topline. On the local pitch event level comprising
      accents and boundary tones the f0 shape is represented by a
      third-order polynomial (left). It's Gestalt properties, i.e. its
      register deviation from the phrase-level register is quantified
      by generating a local register representation the same way as
      for the phrase level (right) and by calculating the root mean
      squared deviations between the midlines and the range regression
      lines.}
    \label{fig:superpos}
  \end{center}
\end{figure}

Next to general f0 and energy features we derived register and local
pitch event related features from the contour-based, parametric, and
superpositional CoPaSul stylization framework \cite{ReichelCSL2014}
representing f0 as a superposition of a global register and a local
pitch accent component. This stylization is presented in Figure
\ref{fig:superpos}. Furthermore rhythmic features were extracted as
described below. All features introduced here as well as the automatic
extraction of prosodic structure can be carried out by means of the
open source CoPaSul prosody analyses software \cite{ReichelArxiv2016,
  copasulGithub}.

All features are listed in Table \ref{tab:feat} together with the
feature set name they belong to and a short description. A more
detailed description is given in the subsequent sections.

\begin{center}
  \begin{table}
    \begin{tabular}{lll}
      \hline
      Feature set & Feature & Description \\
      \hline
      gnl\_en & max & energy maximum in turn\\
      gnl\_en & med & energy median in turn\\
      gnl\_en & sd & energy standard deviation in turn\\
      \hline
      gnl\_f0 & max & f0 maximum in turn\\
      gnl\_f0 & med & f0 median in turn\\
      gnl\_f0 & sd & f0 standard deviation in turn\\
      \hline
      phrase & rng.c0.F/L & f0 range intercept of first/last phrase \\
      phrase & rng.c1.F/L & f0 range slope of first/last phrase \\
      phrase & lev.c0.F/L & f0 level intercept of first/last phrase \\
      phrase & lev.c1.F/L & f0 level slope of first/last phrase \\
      \hline
      acc & c0-3.F/L & polynomial coefficients of the first/last pitch accent \\
      acc & rng.c0.F/L & f0 range intercept of first/last pitch accent \\
      acc & rng.c1.F/L & f0 range slope of first/last pitch accent \\
      acc & lev.c0.F/L & f0 level intercept of first/last pitch accent \\
      acc & lev.c1.F/L & f0 level slope of first/last pitch accent \\
      acc & gst.lev.F/L & level deviation of first/last pitch accent \\
      acc & gst.rng.F/L & range deviation of first/last pitch accent \\
      \hline
      rhy\_en & syl.rate & mean syllable rate \\
      rhy\_en & syl.prop & syllable influence on energy contour \\
      \hline
      rhy\_f0 & syl.prop & syllable influence on f0 contour \\
    \end{tabular}
    \caption{Description of prosodic features grouped by feature
      sets. ``first/last'' refers to the position of the prosodic
      event within a turn.}
    \label{tab:feat}
  \end{table}
\end{center}

\subsection{General f0 and energy features}

For the feature sets {\em gnl\_f0} and {\em gnl\_en} within each turn
we calculated the median, the maximum, and the standard deviation of
the f0 and the energy contour, respectively.

\subsection{Prosodic phrase characteristics}
\label{sec:glob}

The {\em phrase} feature set describes f0 register
characteristics. According to \cite{Rietveld2003} f0 register in the
prosodic phrase domain can be represented in terms of the f0 range
between high and low pitch targets, and the f0 mean level within this
span. To capture both register aspects, level and range, within each
prosodic phrase we fitted a base-, a mid, and a topline by means of
linear regressions as shown in Figure \ref{fig:superpos}. This line
fitting procedure works as follows: A window of length 50 ms is
shifted along the f0 contour with a step size of 10 ms. Within each
window the f0 median is calculated (1) of the values below the 10th
percentile for the baseline, (2) of the values above the 90th
percentile for the topline, and (3) of all values for the
midline. This gives three sequences of medians, one each for the
base-, the mid-, and the topline, respectively. These lines are
subsequently derived by linear regressions, time has been normalized
to the range from 0 to 1. As described in further detail in
\cite{ReichelMadyIS2014} this stylization is less affected by local
events as pitch accents and boundary tones and does not need to rely
on error-prone detection of local maxima and minima. Based on this
stylization the midline is taken as a representation of pitch level.
For pitch range we fitted a further regression line through the
pointwise distances between the topline and the baseline. A negative
slope thus indicates convergence of top- and baseline, whereas a
positive slope indicates divergence.

From this register level and range representation we extracted for the
first and for the (occasionally identical) last prosodic phrase in a
turn the following features: intercept and slope of the midline, and
intercept and slope of the range regression line. That gives eight
features subsumed to the {\em phrase} feature set.

\subsection{Pitch accent characteristics}
\label{sec:loc}

After subtracting the midline derived on the phrase level as described
in section \ref{sec:glob} we fitted third-order polynomials to the
residual f0 contour around the syllable nuclei associated with the
first and the last local pitch event (accent or boundary tone) in a
turn. The stylization window of length 300 ms was placed symmetrically
on the syllable nucleus, and time $t$ was normalized to the range from
-1 to 1. This window length of approximately 1.5 syllables was chosen
to capture the f0 contour on the accented syllable in some local
context.

\begin{figure}
  \begin{center}
    \includegraphics[width=8cm]{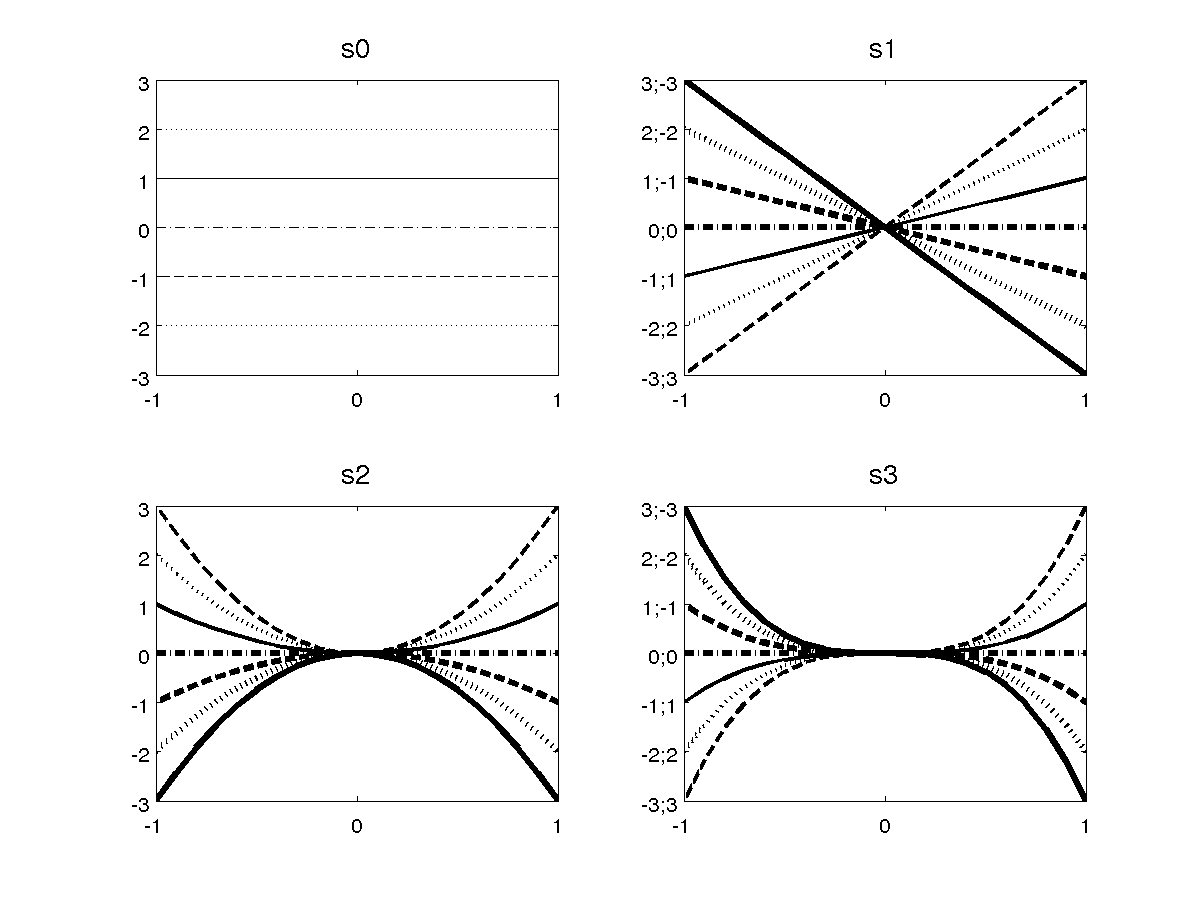}
    \caption{Influence of each coefficient of the third order
      polynomial $\sum_{i=0}^3 s_i\cdot t^i$ on the contour shape. All
      other coefficients set to 0.  For compactness purpose on the
      y-axis both function and coefficient values are shown if they
      differ.}
    \label{fig:poly}
  \end{center}
\end{figure}

As can be seen in Figure \ref{fig:poly} the coefficients represent
different aspects of local f0 shapes. Given the polynomial
$\sum_{i=0}^3 s_i\cdot t^i$, $s_0$ is related to the local f0 level
relative to the register midline.  $s_1$ and $s_3$ are related to the
local f0 trend (rising or falling) and to peak alignment. $s_2$
determines the peak curvature (convex or concave) and its acuity.

Next to the polynomial coefficients we measured local register values
by re-applying the stylization introduced in section \ref{sec:glob}
within the analysis window around the pitch accent.

Finally, pitch accent Gestalt was measured in terms of local register
deviation from the corresponding stretch of global register. This was
simply done by calculating the RMSD between the pitch accent midline
and the corresponding part of the phrase midline. For the accent and
phrase range regression lines we did the same.

From these stylizations the feature set {\em acc} emerges for the
first and for the last local pitch event in a turn. It contains (1)
the polynomial coefficients describing the local f0 shape, (2) the
intercept and slope coefficients for the mid- and the range regression
line describing the local register, and (3) the local level and range
deviation from the underlying phrase in terms of the RMSD between the
accent- and phrase-level regression lines.

\subsection{Rhythm features}

In our approach, rhythm within a turn is represented in terms of
syllable rate (number of detected syllable nuclei per second) and the
influence of the syllable level of the prosodic hierarchy on the
energy and f0 contours. To quantify the syllabic influence on any of
these contours we performed a discrete cosine transform (DCT) on this
contour as in \cite{HeinrichJASA2014}. We then calculated the syllable
influence $w$ as the relative weight of the coefficients around the
syllable rate $r$ ($+/- 1$ Hz to account for syllable rate
fluctuations) within all coefficients below 10 Hz as follows:

\begin{eqnarray}
w & = & \frac{\sum_{c: r-1 \leq f(c) \leq r+1 \textrm{Hz}} |c|}{\sum_{c: f(c)\leq 10 \textrm{Hz}} |c|} \nonumber
\end{eqnarray}

\begin{figure}
  \begin{center}
    \includegraphics[width=8cm]{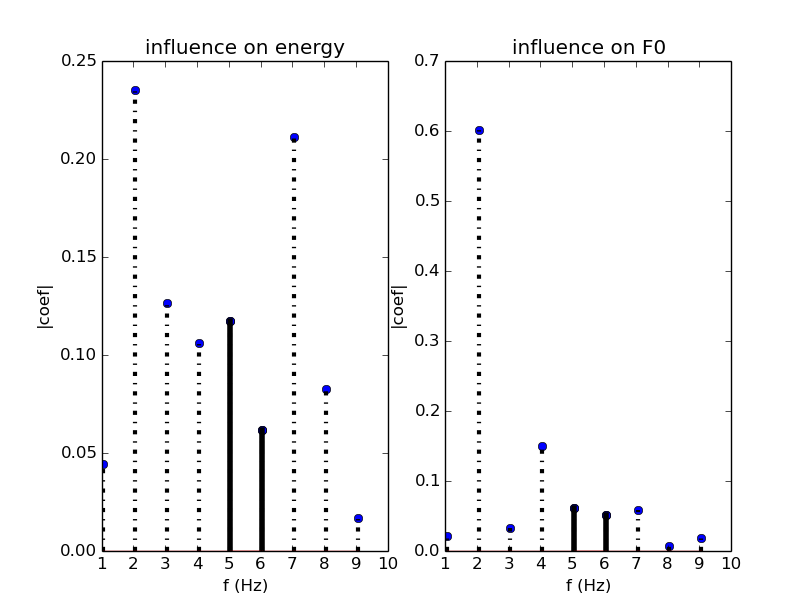}
    \caption{Rhythm features: Quantifying the influence of syllable
      rate on the energy and f0 contour. For this purpose a discrete
      cosine transform (DCT) is applied to the contour. The absolute
      amplitudes of the coefficients around the syllable rate are
      summed and divided by the summed absolute amplitudes of all
      coefficients below 10 Hz. This gives the proportional influence
      of the syllable on the contour. In the shown case the syllable
      rate of 4.5 Hz has a relatively high impact on the energy
      contour but not on the f0 contour. For both contours a high
      influence in the 2 Hz region related to pitch accents can be
      observed.}
    \label{fig:rhy}
  \end{center}
\end{figure}

The higher $w$ the higher thus the influence of the syllable rate on
the contour. This procedure which is shown in Figure \ref{fig:rhy} was
first used to quantify the impact of hand stroke rate on the energy
contour in counting out rhymes \cite{FR_PP16}. The upper cutoff of 10
Hz goes back to the reasoning that contour modulations above 10 Hz do
not occur due to macroprosodic events as accents or syllables, but
amongst others due to microprosodic effects.

\section{Entrainment profiles}
\label{sec:prof}

For all feature sets described in the previous section we generated
entrainment profiles that document in how far speakers entrain with
respect to these features and depending on the speaker's gender and
role in the dialog.%
\footnote{The usage of such profiles was inspired by the speaker
  profile study of \cite{Niebuhr2016}.}

Entrainment generally is expressed in low feature distances relative
to a reference. We address two types of feature distance, one related
to proximity, the other to synchrony. Additionally, we examine
entrainment on a local and a global level based on an asymmetric
pairing of turns to tease apart the impact by speaker genders and
roles. We describe these operationalizations of entrainment in two
subsections \ref{sec:convsync} and \ref{sec:locglob} below and then
proceed to describing the profiles themselves as a means to visualize
the data and generate hypotheses on entrainment for further
statistical testing.

\subsection{Proximity- vs. synchrony-related distance}
\label{sec:convsync}

\begin{figure}
  \begin{center}
    \includegraphics[width=12cm]{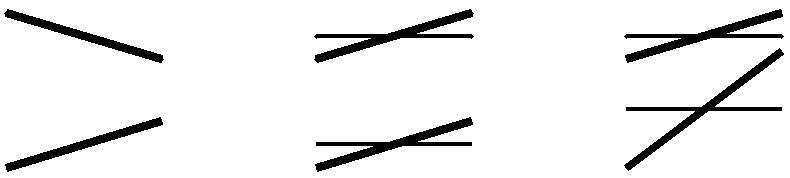}
    \caption{Convergence (left) vs. synchrony (mid)
      vs. convergence+synchrony (right) of some feature.  Convergence
      describes an increase in proximity, which is given for a low
      absolute distance of the feature values. For synchrony the
      feature values are centered on the speaker-dependent mean value
      before calculating their absolute distance.}
    \label{fig:dist}
  \end{center}
\end{figure}

As pointed out in \cite{Edlund2009, Levitan2011, LoozeSC2014}
accommodation can be expressed, amongst others, in terms of proximity
(or similarity), convergence and synchrony. Convergence and proximity
are linked in a way that the former describes an increase of the
latter and thus a decrease in distance over time, which is visualized
in Figure \ref{fig:dist}. Convergence- and proximity-related distance
is trivially represented by a low absolute distance of the feature value
pair, the lower the distance, the higher the proximity. In the
following we restrict the analysis to proximity, thus we are measuring
pointwise distances of single turn pairs without their time
course. Synchrony means that feature values move in
parallel. \cite{LoozeSC2014} proposes to calculate correlations over a
sequence of turn pairs. Here as for proximity we choose a more
straight-forward approach operating on a single turn-pair only. We
simply subtract the respective speakers' mean values from the feature
values before calculating the absolute distance. Synchrony-related
distance is thus low, if the speakers realize a feature either both
above or below their respective means. By that we derive for each
feature and each turn pair one proximity- and one synchrony-related
distance value. It is likely that some of the examined features
preferably undergo one entrainment type only. Pitch accent shape
coefficients for example cannot simply be shifted in parallel by the
interlocutors due to non-linearities in f0 contour continua as found
e.g. by \cite{Kohler1987}, so that for these parameters entrainment is
expected rather to happen not in terms of synchrony but of proximity.

\subsection{Directed local vs. global entrainment}
\label{sec:locglob}

Turn pairing was carried out on two levels to account for local and
for global entrainment. Local entrainment refers to a greater
similarity in adjacent compared to non-adjacent turns in the same
dialog. Global entrainment refers to an overall greater similarity
within a speaker pair (or dialog) than across dyads (dialogs)
\cite{Levitan2011}. For {\bf local entrainment} we compared the
feature distances between adjacent and non-adjacent turns within the
same dialog and task. For the {\em adjacent} sample we paired each
turn with the one directly preceding it in the dialog. For the {\em
  non-adjacent} sample for each turn a non-adjacent turn was drawn
randomly (if available) from the preceding part of the dialog within
the same task among those turns that fulfill the constraint of a
minimum inter-onset interval of 15 seconds. For {\bf global
  entrainment} feature distances were compared between turn pairs in
the same dialog and the same number of turn pairs across dialogs. For
the {\em same dialog} sample we paired each turn with a randomly drawn
turn from the preceding part of the same dialog and task. For the {\em
  different dialog} sample we randomly paired turns of unrelated
speaker pairs, i.e. speaker pairs not engaged in any common game
conversation.

Our sample generation approach differs from previous approaches as in
\cite{Levitan2011} in several respects: first, we apply a directed
pairing of turns to the left dialog context only. This enables us to
compare entrainment behavior asymmetrically across speaker genders and
roles since for the statistical analyses described below we relate the
obtained distance values not to both speakers but to the second one
only, i.e. for each turn pair we examine how similar the second
speaker gets to the first, and not vice versa.

Second, for global entrainment we are not comparing mean feature
values calculated for each speaker as \cite{Levitan2011}, but
analogously as for local entrainment we work on the raw turn pair
data. This ensures comparable sample sizes in local and global
entrainment examination making the results less dependent on the
number of speakers in the corpus, especially if this number is
low. And again this approach allows for asymmetric examination of
gender and role influence also on global entrainment. As opposed to
mean value comparison that yields one distance value for both speakers
in a dialog, directed turn comparison assigns a distinct value to each
interlocutor.

Third, our approach differs with respect to IPU pairing. {\em
  Adjacency} refers to the turn level and not to the compared events
themselves. For each turn pair we compared separately their initial
and their final phrase and accent characteristics, which implies that
also for adjacent turns the compared events generally are not
adjacent. This approach is motivated first by our goal to compare
entrainment effects in dependence of the position within a
turn. Furthermore, it serves to reduce value range differences across
different positions within an utterance. These differences are amongst
others caused by declination and locally restricted event functions
such as pitch accents vs. boundary tones. By our positional
restriction we obtain distance values with a less obscured link to
entrainment.

\subsection{Profiles}

\begin{figure}
  \begin{center}
    \includegraphics[width=12.6cm]{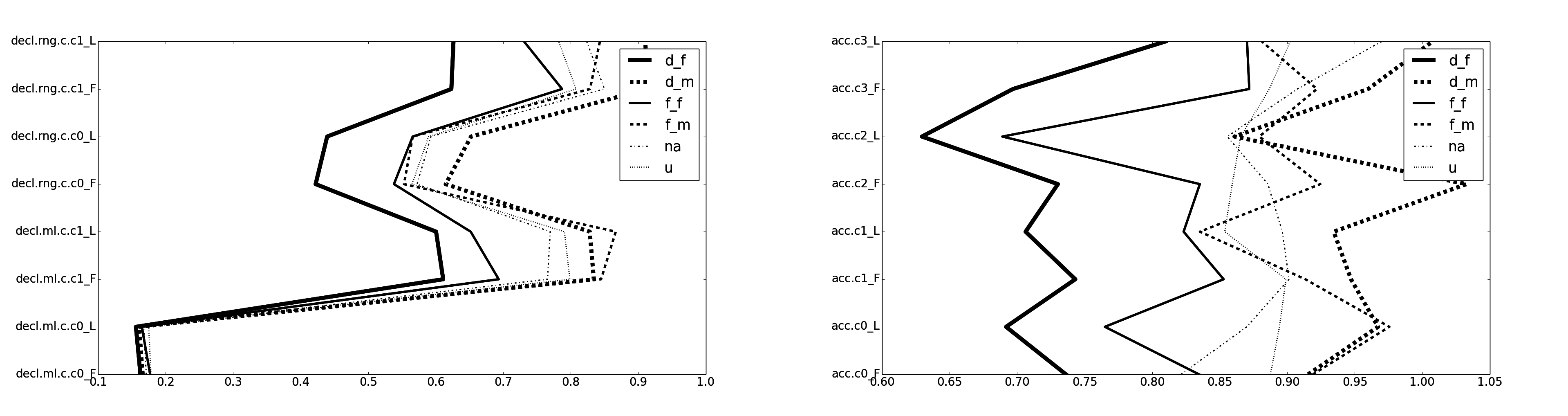}
    \caption{Entrainment profiles for features from the sets {\em
        phrase} (left) and {\em acc} (right). The y-axis gives the
      features described in table \ref{tab:feat}, the x-axis gives
      their mean proximity-related distances. For each speaker type
      {\em role\_gender} defined by role (describer {\em d} or
      follower {\em f}) and gender (female {\em f} or male {\em m}) a
      profile graph relates each feature to its mean proximity
      distance in adjacent turns. Describers {\em d\_*} profiles are
      given in thick lines, follower {\em f\_*} profiles in thin
      lines. Solid indicates female {\em *\_f}, dashed male {\em
        *\_m}. Two reference profiles are given for non-adjacent turns
      in the same dialog ({\em na}, dash-dotted) and for unrelated
      turns in different dialogs ({\em u}, dotted).}
    \label{fig:prof_copa}
  \end{center}
\end{figure}

For each feature set we generated for proximity and for synchrony each
an entrainment profile in the following way: the features of the
respective set are plotted on the y axis, and their mean distance
values on the x axis. Mean distance values are separately calculated
for adjacent turns ({\em a}), non-adjacent turns in the same dialog
({\em na}) and turn pairs across dialogs ({\em u}). The latter two
define the references for local and global entrainment,
respectively. The adjacent turn distances are further split by speaker
role and gender, to visualize the impact of speaker type on
entrainment. Distance and type specification always refers to the
responding speaker, i.e. the speaker uttering the later turn in the
turn pair. For visual inspection a local entrainment tendency is
indicated by {\em a}-lines left of the {\em na}-reference line. Global
entrainment is reflected in {\em a}-lines and the {\em na}-line left
of the {\em u}-reference line. For both the local and global domain
the opposite order indicates a disentrainment tendency. Figure
\ref{fig:prof_copa} shows mean proximity distance values for the
feature sets {\em phrase} and {\em acc}. By visual inspection female
speakers (solid lines), especially the followers (thick solid) show
smaller distances in adjacent turns than in non-adjacent or unrelated
ones for most features indicating entrainment. Male speaker profiles
(dashed lines), especially the describer ones (thick dashed), in
contrast are right of the reference lines indicating higher distance
values and thus a disentrainment tendency.

\begin{figure}
  \begin{center}
    \includegraphics[width=12.6cm]{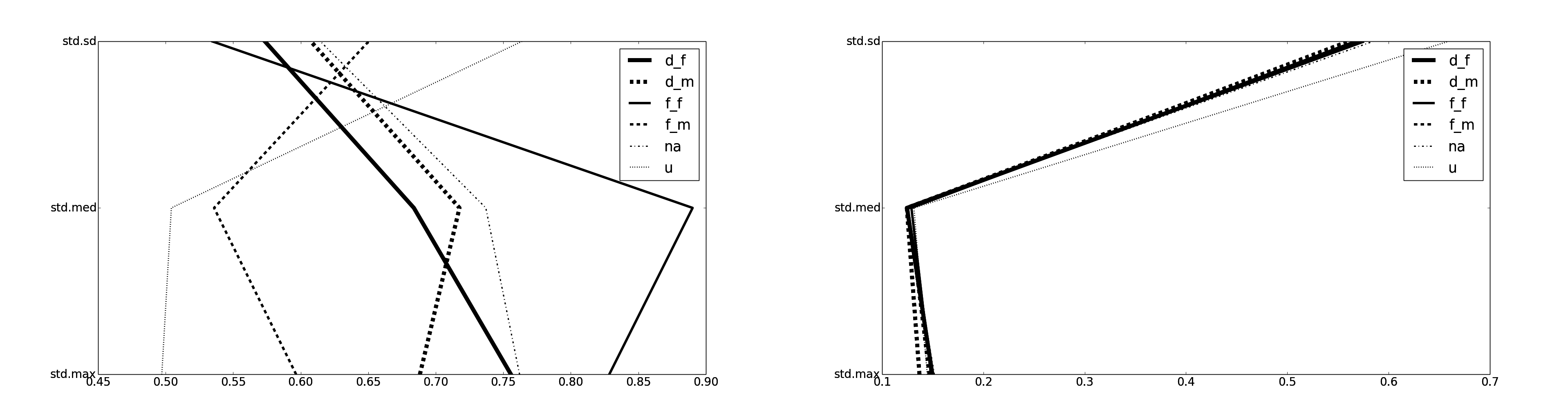}
    \caption{Entrainment profiles for the feature set {\em gnl\_f0}
      for proximity (left) and synchrony (right). For this feature set
      these two entrainment measures behave very differently. For
      details please see the caption of Figure \ref{fig:prof_copa}.}
    \label{fig:prof_gnl}
  \end{center}
\end{figure}

\subsection{Descriptive observation}

By visual inspection of such entrainment profiles in Figures
\ref{fig:prof_copa} and \ref{fig:prof_gnl} the following observations
can be made:

\begin{itemize}
\item There is a role-gender interaction; female describers (thick
  solid) generally entrain most, male describers (thick dashed)
  entrain least.
\item The zigzag lines for entraining speakers in Figure
  \ref{fig:prof_copa} for set {\em acc} indicate that more entrainment
  takes place in turn-final than in turn-initial position ({\em *\_L}
  and {\em *\_F} features, respectively).
\item The feature sets undergo entrainment to a different
  degree. While {\em gnl\_en} features do not entrain at all, {\em
    acc} features show entrainment for certain speaker types.
\item The profile pair in Figure \ref{fig:prof_gnl} suggests that there is
  a bias of some feature sets towards proximity or synchrony.
\end{itemize}

These descriptive observations obtained from the visualization of
entrainment profiles serve as hypotheses for further statistic
examinations that are described in the following section.

\section{Harvesting and condensation of entrainment data}
\label{sec:harv}

To cope with the complexity of our data -- 37 acoustic features times
2 entrainment domains times 2 distance measures times each 2 roles and
genders -- we employed a two-step approach consisting of data
harvesting and condensation. By harvesting we collect the entrainment
behavior of all speaker types for all prosodic features. Subsequent
condensation serves to structure the data in terms of probabilistic
relations between entrainment on one hand, and feature sets, speaker
types, and segment positions on the other hand.

\subsection{Harvesting}
\subsubsection{Methods}

We used linear mixed-effect models for each prosodic feature based on
the lmer() function in the {\em lme4} package in the statistics
software R \cite{lme4_bates2015}.  The dependent variable {\em dist}
refers to proximity and synchrony each in global and local entrainment
turn pairs. Thus for each prosodic feature, 4 distance values are
tested. The fixed effects are {\em pairing, role}, and {\em
  gender}. For local entrainment {\em pairing} stands for {\em
  adjacent vs. non-adjacent} turn. For global entrainment it stands
for {\em same vs. different} dialog. {\em role} and {\em gender} refer
to the replying speaker and define his/her role in the play ({\em
  describer} or {\em follower}) and the gender ({\em female} vs. {\em
  male}). The identities of the initiating and the replying speaker
are considered to be random factors for which a random intercept model
was calculated. Significant interactions ($p<0.05$) of the fixed
effects calculated by the Anova() function of the \emph{car} package
in R \cite{car_package2011} were subsequently examined by re-applying
the tests on corresponding subsets. To account for the large number of
tests, $p$-values were corrected for false discovery rate
\cite{Benjamini2001}.

\subsubsection{Results}
\label{sec:harvres}
From these tests we derived two tables \ref{tab_glb_rawCompact} and
\ref{tab_loc_rawCompact} for global and local entrainment,
respectively.

\begin{table}
\centering
\begingroup\tiny
\begin{tabular}{rll|ll|ll}
  \hline
 & \multicolumn{2}{l|}{Features} & \multicolumn{2}{l|}{Entrainment} & \multicolumn{2}{l}{Disentrainment} \\
 & set & name & prox & sync & --prox & --sync \\ 
  \hline
  1 & gnl\_en & max & -- & -- & x\_x,x\_m,d\_m,f\_m & x\_x,x\_m,d\_m,f\_m \\ 
  2 & gnl\_en & med & -- & -- & -- & -- \\ 
  3 & gnl\_en & sd & -- & x\_f & x\_m,d\_m,f\_m & x\_m,d\_m,f\_m \\ 
  4 & gnl\_f0 & max & -- & f\_f & x\_x,x\_f,x\_m,d\_x, & d\_f \\ 
     &         &     &    &      & d\_f,d\_m,f\_x,f\_f,f\_m & \\
  5 & gnl\_f0 & med & f\_m & x\_x & x\_x,x\_f,x\_m,d\_f, & -- \\
     &         &     &      &      & d\_m,f\_f & \\
  6 & gnl\_f0 & sd & x\_x,x\_f,x\_m,d\_f, & x\_x,f\_f & -- & -- \\ 
     &         &    & d\_m,f\_f,f\_m & & & \\
  7 & phrase & lev.c0.F & -- & x\_x,d\_m & x\_x,x\_f,x\_m,d\_x,f\_x & -- \\ 
    &      &              &    & f\_x & \\
  8 & phrase & lev.c0.L & f\_m & x\_x & x\_x,x\_f,x\_m,d\_x, & -- \\
    &      &              &      &      & d\_f,d\_m,f\_x,f\_f            & \\ 
  9 & phrase & lev.c1.F & x\_x & x\_x & -- & -- \\ 
  10 & phrase & lev.c1.L & x\_x & -- & -- & -- \\ 
  11 & phrase & rng.c0.F & x\_x,x\_f,d\_f & d\_f & -- & d\_m \\ 
  12 & phrase & rng.c0.L & x\_x,x\_f,d\_f,f\_f & d\_f,f\_f & -- & d\_m \\ 
  13 & phrase & rng.c1.F & d\_f & d\_f & x\_m,d\_m,f\_m & x\_m,d\_m,f\_m \\ 
  14 & phrase & rng.c1.L & d\_f & d\_f & x\_m,d\_m & d\_m \\ 
  15 & acc & c0.F & x\_x & -- & -- & -- \\ 
  16 & acc & c0.L & x\_x & x\_x,x\_f & -- & -- \\ 
  17 & acc & c1.F & -- & -- & x\_m & x\_m \\ 
  18 & acc & c1.L & -- & -- & -- & -- \\ 
  19 & acc & c2.F & d\_f & d\_f & x\_m & x\_m,d\_m \\ 
  20 & acc & c2.L & x\_x,d\_f & x\_x,d\_f & -- & -- \\ 
  21 & acc & c3.F & d\_f & d\_f & d\_m,f\_m & d\_m,f\_m \\ 
  22 & acc & c3.L & -- & -- & -- & -- \\ 
  23 & acc & lev.c0.F & f\_m & x\_x & x\_x,x\_f,x\_m,d\_f,d\_m, & -- \\ 
     &     &              &      &      & f\_f & \\
  24 & acc & lev.c0.L & f\_m & x\_x & x\_x,x\_f,x\_m,d\_m,f\_f & -- \\ 
  25 & acc & lev.c1.F & -- & -- & x\_m,d\_m & x\_m,d\_m \\ 
  26 & acc & lev.c1.L & x\_x & x\_x & -- & -- \\ 
  27 & acc & rng.c0.F & x\_x,x\_f,d\_f & -- & x\_m,f\_m & x\_m \\ 
  28 & acc & rng.c0.L & x\_x,d\_f & -- & -- & -- \\ 
  29 & acc & rng.c1.F & d\_f & d\_f & x\_m,d\_m & x\_m,d\_m \\ 
  30 & acc & rng.c1.L & -- & -- & x\_m,d\_m,f\_m & x\_m,d\_m,f\_m \\ 
  31 & acc & gst.lev.rms.F & x\_f & d\_f & x\_m & x\_m,d\_m \\ 
  32 & acc & gst.lev.rms.L & x\_x,x\_f,d\_x & -- & -- & f\_m \\ 
  33 & acc & gst.rng.rms.F & x\_f & -- & -- & f\_m \\ 
  34 & acc & gst.rng.rms.L & x\_x,x\_f & -- & -- & -- \\ 
  35 & rhy\_en & syl.prop & -- & -- & -- & -- \\ 
  36 & rhy\_en & syl.rate & -- & -- & -- & f\_f \\ 
  37 & rhy\_f0 & syl.prop & x\_f & -- & -- & -- \\ 
   \hline
\end{tabular}
\endgroup
\caption{Global entrainment and disentrainment by feature and speaker
  type for proximity {\em prox} and synchrony {\em sync}. Speaker type
  is encoded as {\em role\_gender}; role: describer {\em d}
  vs. follower {\em f}; gender: female {\em f} vs. male {\em m}; {\em
    x} denotes {\em not specified}. To give an example how to read
  this table: line 14 refers to the feature {\em rng.c1.L} of the {\em
    phrase} set, i.e. the range slope of the turn-final phrase. For
  this feature female describers {\em d\_f} entrain with respect to
  both proximity and synchrony. Proximity disentrainment is observed
  for male speakers {\em x\_m} which turned out to be significant due
  to the disentraining behavior of male describers {\em d\_m}.}
\label{tab_glb_rawCompact}
\end{table}

\begin{table}
\centering
\begingroup\tiny
\begin{tabular}{rll|ll|ll}
  \hline
 & \multicolumn{2}{l|}{Features} & \multicolumn{2}{l|}{Entrainment} & \multicolumn{2}{l}{Disentrainment} \\
 & set & name & prox & sync & --prox & --sync \\ 
  \hline
  1 & gnl\_en & max & -- & -- & x\_x,d\_f,d\_m,f\_f & x\_x,d\_f,d\_m,f\_f \\ 
  2 & gnl\_en & med & -- & -- & x\_x & x\_x \\ 
  3 & gnl\_en & sd & -- & -- & x\_x,d\_x,d\_f,d\_m, & x\_x,d\_x,d\_f,d\_m, \\
     &         &    &    &    & f\_x,f\_f,f\_m & f\_x,f\_f,f\_m \\
  4 & gnl\_f0 & max & -- & d\_x,d\_f & -- & -- \\ 
  5 & gnl\_f0 & med & -- & d\_x & -- & -- \\ 
  6 & gnl\_f0 & sd & x\_x,d\_f & d\_f,f\_m & -- & -- \\ 
  7 & phrase & lev.c0.F & f\_f & -- & -- & -- \\ 
  8 & phrase & lev.c0.L & x\_f & -- & -- & -- \\ 
  9 & phrase & lev.c1.F & d\_x & d\_x & -- & -- \\ 
  10 & phrase & lev.c1.L & d\_x & d\_x & -- & -- \\ 
  11 & phrase & rng.c0.F & -- & -- & -- & -- \\ 
  12 & phrase & rng.c0.L & -- & d\_f & -- & -- \\ 
  13 & phrase & rng.c1.F & -- & x\_x,x\_m,f\_m & -- & -- \\ 
  14 & phrase & rng.c1.L & -- & -- & -- & -- \\ 
  15 & acc & c0.F & -- & -- & -- & -- \\ 
  16 & acc & c0.L & f\_x & -- & -- & -- \\ 
  17 & acc & c1.F & -- & -- & -- & -- \\ 
  18 & acc & c1.L & -- & -- & -- & -- \\ 
  19 & acc & c2.F & -- & -- & -- & -- \\ 
  20 & acc & c2.L & x\_x,f\_f & x\_x,f\_f & -- & -- \\ 
  21 & acc & c3.F & -- & -- & -- & -- \\ 
  22 & acc & c3.L & x\_x & x\_x & -- & -- \\ 
  23 & acc & lev.c0.F & -- & -- & -- & -- \\ 
  24 & acc & lev.c0.L & -- & -- & -- & -- \\ 
  25 & acc & lev.c1.F & -- & -- & -- & -- \\ 
  26 & acc & lev.c1.L & -- & -- & -- & -- \\ 
  27 & acc & rng.c0.F & -- & -- & -- & -- \\ 
  28 & acc & rng.c0.L & -- & -- & -- & -- \\ 
  29 & acc & rng.c1.F & -- & -- & -- & -- \\ 
  30 & acc & rng.c1.L & -- & -- & -- & -- \\ 
  31 & acc & gst.lev.rms.F & -- & -- & -- & -- \\ 
  32 & acc & gst.lev.rms.L & -- & -- & -- & -- \\ 
  33 & acc & gst.rng.rms.F & -- & -- & x\_x & -- \\ 
  34 & acc & gst.rng.rms.L & -- & -- & -- & -- \\ 
  35 & rhy\_en & syl.prop & x\_x,d\_f,d\_m & -- & -- & -- \\ 
  36 & rhy\_en & syl.rate & x\_x,d\_f,f\_m & x\_x,d\_m & -- & -- \\ 
  37 & rhy\_f0 & syl.prop & -- & -- & -- & -- \\ 
   \hline
\end{tabular}
\endgroup
\caption{Local entrainment and disentrainment by feature and speaker
  type for proximity {\em prox} and synchrony {\em sync}. Speaker type
  is encoded as {\em role\_gender}; role: describer {\em d}
  vs. follower {\em f}; gender: female {\em f} vs. male {\em m}; {\em
    x} denotes {\em not specified}. To give an example how to read
  this table: line 16 refers to the feature {\em c0.L} of the {\em
    acc} feature set, i.e. the coefficient $c_0$ of the polynomial
  stylization of the turn final local pitch event. For this feature
  all followers {\em f\_x} entrain with respect to proximity.}
\label{tab_loc_rawCompact}
\end{table}

In Tables \ref{tab_glb_rawCompact} and \ref{tab_loc_rawCompact} the
columns {\em prox} and {\em sync} contain all speaker types for which
the linear mixed-effect models introduced in the previous section
revealed entrainment for a certain feature and distance measure
($\alpha=0.05$, $p$-values corrected for false discovery
rate). Speaker types are composed of the speaker's role (describer
{\em d} vs. follower {\em f}), and gender (female {\em f} vs. male
{\em m}). For local entrainment this means, that the distance of a
feature is significantly smaller in neighboring turns opposed to
non-neighboring turns. For global entrainment it indicates, that the
distance is significantly smaller within a dialog than across
dialogs. The {\em --prox} and {\em --sync} columns show all
disentraining speaker types for a feature and a distance measure, that
is, for adjacent or within-dialog turn pairs the distance turned out
to be significantly higher than for non-adjacent/cross-dialog turn
pairs.

\subsection{Condensation}
\subsubsection{Method}

From the tables obtained by harvesting we infer conditional
entrainment probabilities separately for proximity and synchrony for
feature sets, position within a turn, and speaker type as exemplified
for the feature set {\em gnl\_f0} and proximity. In Table
\ref{tab_loc_rawCompact} in one out of three cases (row 6 out of 4--6)
column {\em prox} reports entrainment evidence, which is defined by
the observation that at least one of the speaker types (x\_x and d\_f
in row 6) shows entrainment. Thus the conditional proximity
entrainment probability for feature set {\em gnl\_f0} amounts to
$\frac{1}{3}$. For synchrony entrainment occurs for all features, thus
conditional synchrony entrainment is $1$. Analogously, given no
disentrainment evidence, the disentrainment probability is $0$ both
for proximity and synchrony.

\subsubsection{Results}
\label{sec:res}
\begin{figure}
  \begin{center}
    \includegraphics[width=12cm]{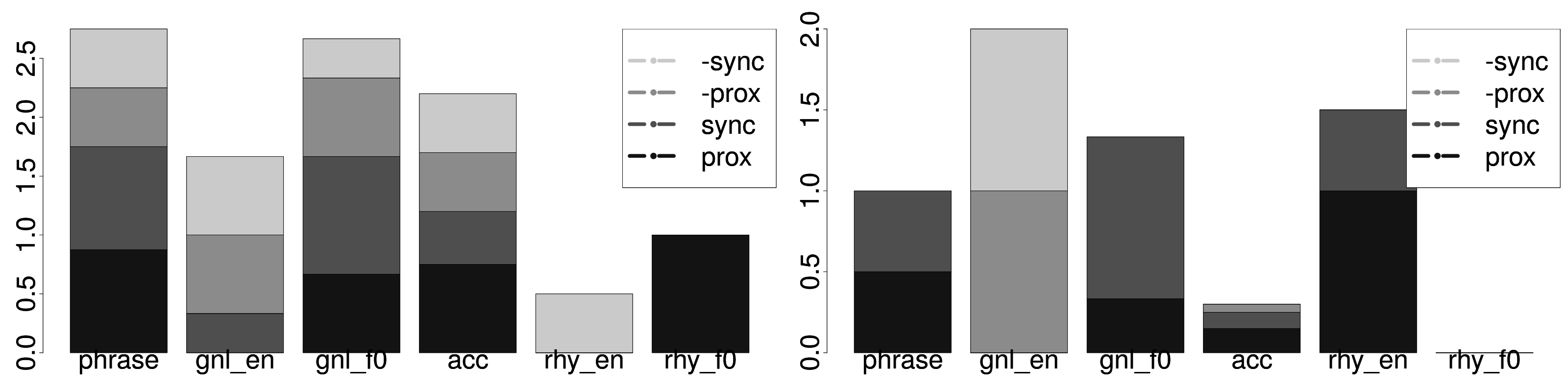}
    \caption{Conditional global (left) and local (right) entrainment
      and disentrainment probabilities for each feature set derived
      from Tables \ref{tab_glb_rawCompact} and
      \ref{tab_loc_rawCompact}. Disentrainment for proximity and
      synchrony is denoted by {\em --prox} and {\em --sync},
      respectively. Each partition in the stacks denotes a probability
      with values between 0 and 1.}
    \label{fig:prob_fset}
  \end{center}
\end{figure}

\begin{figure}
  \begin{center}
    \includegraphics[width=12cm]{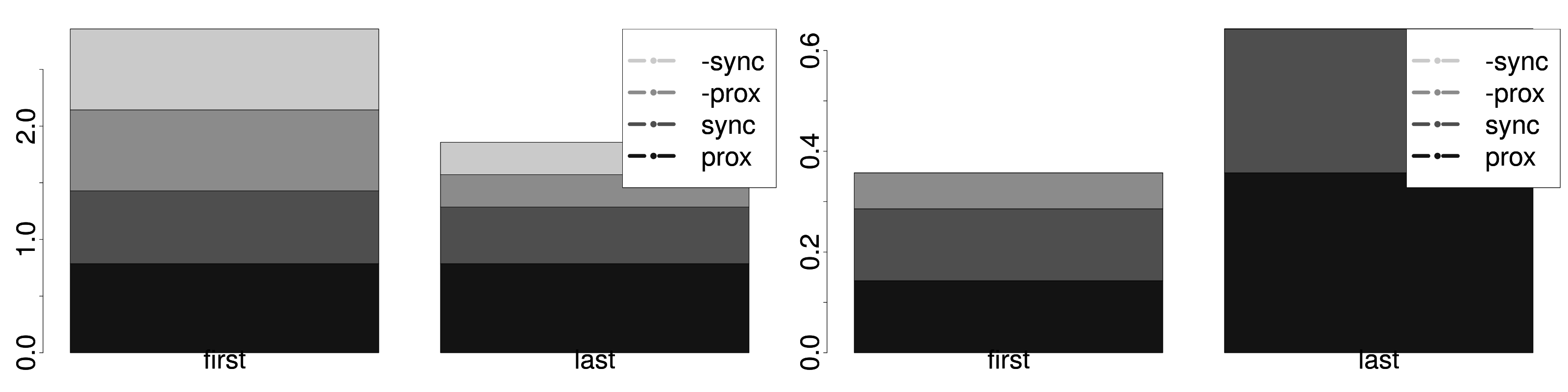}
    \caption{Conditional global (left) and local (right) entrainment
      and disentrainment probabilities for the first and last position
      within turns derived from Tables \ref{tab_glb_rawCompact} and
      \ref{tab_loc_rawCompact}. Disentrainment for proximity and
      synchrony is denoted by {\em --prox} and {\em --sync},
      respectively. Each partition in the stacks denotes a probability
      with values between 0 and 1.}
    \label{fig:prob_pos}
  \end{center}
\end{figure}

\begin{figure}
  \begin{center}
    \includegraphics[width=12cm]{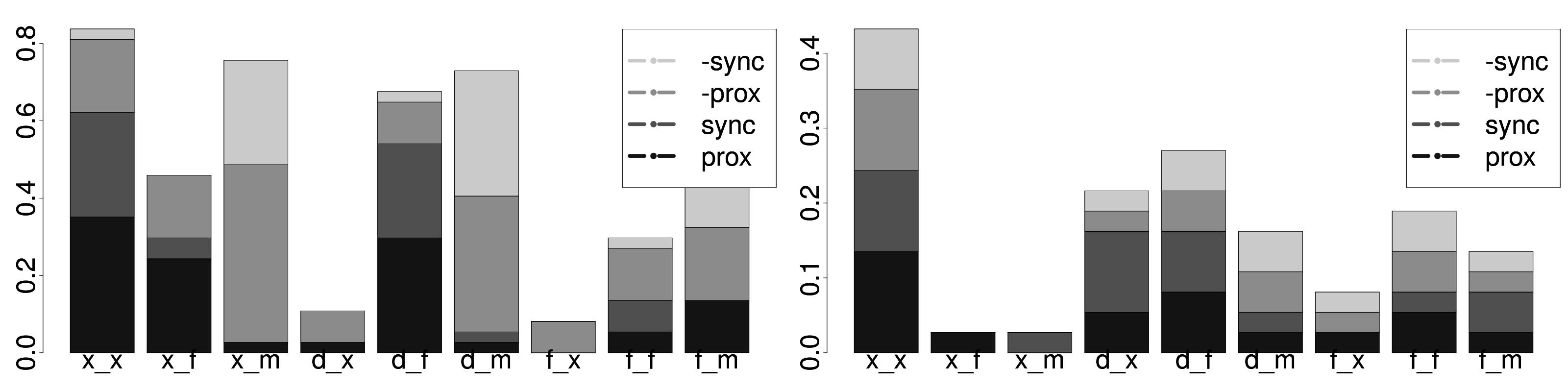}
    \caption{Conditional global (left) and local (right) entrainment
      and disentrainment probabilities derived from Tables
      \ref{tab_glb_rawCompact} and \ref{tab_loc_rawCompact} for each
      speaker type defined by {\em role\_gender}; role: describer {\em
        d} vs. follower {\em f}; gender: female {\em f} vs. male {\em
        m}; {\em x} denotes {\em not specified}. Disentrainment for
      proximity and synchrony is denoted by {\em --prox} and {\em
        --sync}, respectively. Each partition in the stacks denotes a
      probability with values between 0 and 1.}
    \label{fig:prob_spk}
  \end{center}
\end{figure}

Tables \ref{tab_glb_rawCompact} and \ref{tab_loc_rawCompact} show
which speaker types entrain or disentrain in terms of proximity or
synchrony for each feature.  The features are further categorized into
feature sets. In Table \ref{tab_glb_rawCompact} the global entrainment
data is collected, in Table \ref{tab_loc_rawCompact} the local
one. Position of the compared segments within the turns is indicated
in the column {\em feat} by the final capital letters $F$ and $L$ (for
first and last segment, respectively). This categorization only
applies to the feature sets {\em phrase} and {\em acc}. The
conditional probabilities for feature sets, position in the turn, and
speaker types which were derived from the tables as described in
section \ref{sec:harvres} are visualized by stacked barplots in
Figures \ref{fig:prob_fset}, \ref{fig:prob_pos}, and
\ref{fig:prob_spk}.

From the barplots we infer the following observations that will be
discussed in section \ref{sec:disc}:

\begin{enumerate}
\item The feature sets {\em gnl\_f0}, {\em phrase} and {\em acc} show
  strong entrainment tendencies, so that especially the newly
  introduced features of {\em phrase} and {\em acc} are worth to be
  looked at more closely. In contrast, feature set {\em gnl\_en} is
  very much biased towards disentrainment.
\item The new feature sets undergo local and global entrainment to
  different proportions. While {\em phrase} and {\em acc} undergo more
  global entrainment, the opposite is to be observed for {\em
    rhy\_en}.
\item Some feature sets such as {\em acc} tend to show proximity
  whereas other feature sets as {\em gnl\_f0} tend to show
  synchronization.
\item Entrainment takes place in turn-final position more than in
  turn-initial position.
\item Entrainment is highly speaker-type dependent, more precisely
  there is an interaction between role and gender. Female describers
  entrain most, male describers entrain least, female and male
  followers entrain to approximately the same extent.
\end{enumerate}

\subsubsection{Task success}
\begin{figure}
  \begin{center}
    \includegraphics[width=12cm]{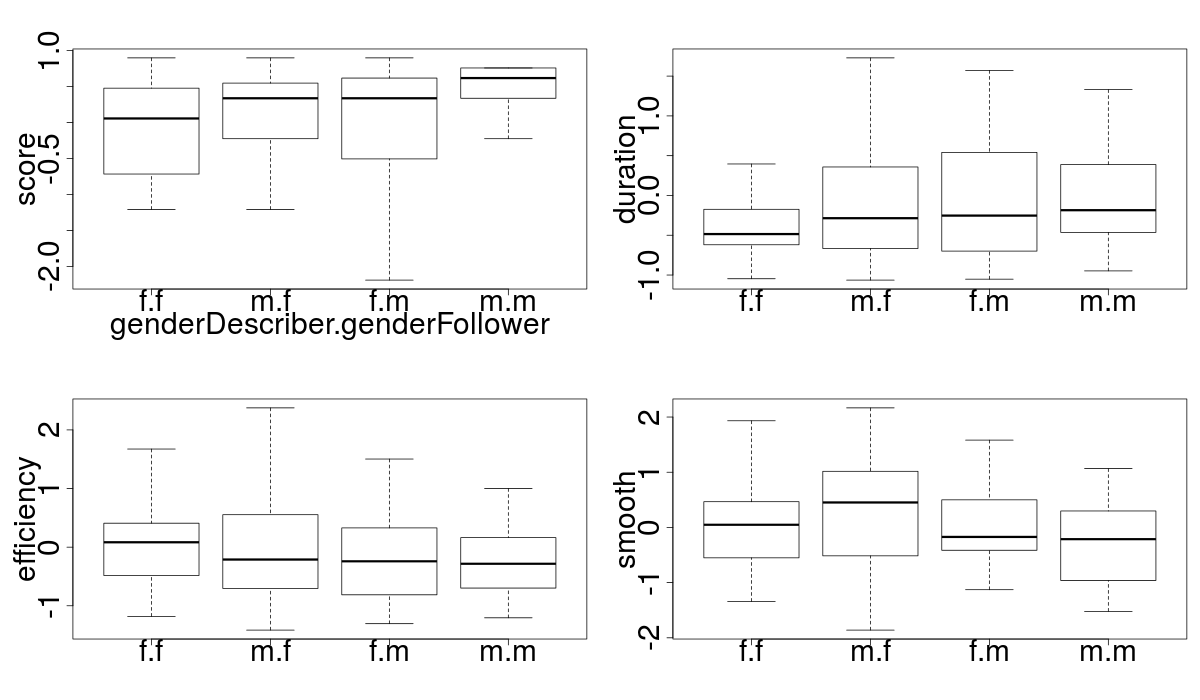}
    \caption{Task success measures (z-transformed) for all gender pairings in the role of describers and followers.}
    \label{fig:score}
  \end{center}
\end{figure}

Next to the entrainment plots we recorded several task success
measures in Figure \ref{fig:score} for all gender/role combinations in
order to examine whether the speaker-type related entrainment behavior
has an impact on task success. {\em Score} measures the distance
between the reference and the game outcome of the target object location as
described in section \ref{sec:data}. {\em Duration} gives the time it
took to solve the task, {\em efficiency} is {\em score} divided by
{\em duration}, and {\em smooth} stands for the proportion of smooth
turn transitions in the entire dialog. A transition was defined to be
smooth, if it falls in the interval between $-0.5$ and $0.5$
seconds. Overlap values below $-0.5s$ and delays above $0.5s$ indicate
interruptions and vacillations, respectively. These values were
selected for two reasons. First, turns with minor overlaps or delays
within one or two syllables are commonly perceived as 'smooth' in
high-involvement interactions (\cite{Tannen1994}). Moreover, we also
examined latencies in an almost identical corpus of collaborative
games in English (\cite{GravanoHirschberg2011}) with hand-annotations
of turn-types such as smooth switch, overlap, interruption, or pause
interruption. We found that interruptions were more likely than plain
overlaps for overlaps greater than 350ms, and that pause interruptions
(signaling non-smooth hesitations from the current speaker) were also
more likely than smooth switches with more than 500ms latency.%
\footnote{We thank A. Gravano for providing us with mean latencies for
  turn types in Columbia Games Corpus.}

We tested differences for each of the four success measures by linear
mixed-effect models with task success as the dependent variable, the
describer and follower gender as the independent variable. The Ids of
both speakers were taken as random effects, for which a random
intercept model was calculated.

Only for {\em score} we found a significant difference for which only the
describer's gender is responsible ($t=1.696$, $p=0.0220$;
follower's gender: $t=1.016$, $p=0.1362$; interaction: $t=0.012$,
$p=0.9903$). Pairs with male describers tend to achieve higher
scores. For none of the other success variables any significant
relationship has been found, neither to the describer's or follower's
gender, nor to their interaction ($t<1.02$, $p>0.135$).

\section{Discussion}
\label{sec:disc}

In the following we will elaborate on the different behavior of
feature sets with respect to proximity and synchrony, as well as with
respect to global and local entrainment (see observed tendencies 1--3
in section \ref{sec:res}). Then we discuss reasons for the observation
that entrainment predominantly occurs in turn-final position (tendency
4). Finally, gender-role interactions (tendency 5) in entrainment will
be explained by different gender-related strategies in
solution-oriented collaborative interactions.

\subsection{Feature sets}

\paragraph{What are the general tendencies of set-related entrainment?}
As can be seen in Figure \ref{fig:prob_fset} next to the well-examined
set {\em gnl\_f0} also the new sets {\em phrase} and {\em acc} not yet
examined in previous studies show clear entrainment tendencies showing
that speakers do not only accommodate in terms of coarse but also more
fine-grained local prosodic characteristics.

Clearly, {\em gnl\_en} shows disentrainment, especially for local
comparisons. This might be due to the cooperative dialog situation in
combination with the potential, especially of energy-related
entrainment, to hinder cooperation, e.g. in cases when both speakers
start to raise their voice as in competitive turn taking situations.
From the more general perspective of Causal Attribution Theory
\cite{Heider1958} one interprets other people's behavior with respect
to their intentions and motivations. \cite{Giles1979} argue in the
Conversation Accommodation Theory (CAT) framework, that proximity
might also be considered as negative if the supposed intent has
negative connotations. Such a negatively received accommodation occurs
for example in {\em patronizing communication} \cite{Giles1991,
  Soliz2014}, which can manifest itself in mimicking dialectal
features \cite{Giles1991} and in a slow and less complex speaking
style of young adults when talking to older adults based on negative
age-related stereotypes \cite{Soliz2014}. Applied to our data, a joint
increase in energy might be considered as a negative accommodation
which is mutually interpreted as confrontational, so that speakers
rather diverge on this feature.

This finding also extends previous observations regarding
(dis)entrainment in this corpus. \cite{BenusCI2014} and
\cite{LevitanDD2015} analyzing local and global entrainment in the
data in terms of proximity, convergence, and synchrony and using
different methodological approaches than this paper, found tendencies
for entrainment in intensity that were, nevertheless, stronger than
for other features. On the one hand, this supports the analysis of
intensity as a feature with low-functional load and thus relatively
free to participate in negotiating social relations during the
dialog. On the other hand, the diverging tendencies in the current
and previous results suggest the complex nature of entrainment in
speech and the possibility that the entrainment potential of certain
features within a dataset might be sensitive to different
operationalizations of entrainment in terms of synchrony, convergence,
local and global domains, or units of analysis.

\paragraph{Do sets differ with respect to global vs. local entrainment?}
{\bf More global entrainment} indicates that overall speakers accommodate,
but local linguistic (e.g. sentence type, dialog act, information status)
variation inhibits local accommodation. This can be observed for the
feature sets {\em phrase} and {\em acc} that clearly are affected by
such linguistic parameters. Analogously, disentrainment for such
features systematically occurs only on the global level.

{\em rhy\_en} in contrast can undergo much {\bf more local
  entrainment} since such rhythm features are much less constrained by
linguistic context than {\em phrase} and {\em acc}.

\paragraph{Proximity vs synchrony}
Feature sets show tendencies to undergo entrainment {\bf either in
  terms of proximity or of synchrony}. Features defining f0 shape,
mainly contained in feature set {\em acc}, show similarity to a higher
extent than synchrony. In contrast overall f0 median, maximum, and
standard deviation features from set {\em gnl\_f0} rather synchronize
than become similar (e.g. both speakers deviate in the same direction
from their mean instead of getting closer). This implies that a
mixed-gender conversation does not lead to a mutually approaching f0
mean, i.e. that the female speaker lowers her pitch, while the male
speaker raises it. Rather the speakers accommodate in such a way that
they both use a high or low register relative to their personal
reference, thus they synchronize. Furthermore (as mentioned above),
synchrony does not disentangle entrainment from competition as in
competitive turn taking situations in which both speakers might signal
their interest to keep/get the turn by a relatively high f0 register
\cite{French1983}. For f0 shapes in contrast, synchrony is much less
likely, since speakers cannot simply shift different f0 contours in
parallel due to non-linearities (e.g. early vs late peak
\cite{Kohler1987}). Rather they accommodate to more similar f0 shapes.

For the feature set {\em phrase} {\bf both synchrony and proximity
  apply} to the same extent as is visualized in the right part of
Figure \ref{fig:dist}. This indicates, that the features are varied in
parallel but not to the same degree, i.e. one speaker additionally
becomes similar towards the other. This asymmetric behavior can be
observed predominantly for describers (cf Tables
\ref{tab_glb_rawCompact} and \ref{tab_loc_rawCompact}, columns {\em
  prox, sync}, rows 7--14) and among them rather for females (cf Table
\ref{tab_glb_rawCompact}, rows 7--14).

\subsection{Within-turn position}

Differentiating between turn-initial and turn-final position reveals
an imbalance in local but not in global entrainment (cf Figure
\ref{fig:prob_pos}). Local entrainment is more likely to occur in
turn-final position. Generally speaking, these different amounts of
local and global as well as of turn-initial and -final entrainment
support the notion of hybrid causes for accommodation as proposed by
\cite{Kraljic2008, Schweitzer2014}, cf section \ref{sec:intro}. Next
to automatic priming mechanisms applying throughout the entire turn it
seems that in turn-final position pragmatic goals are an additional
trigger for entrainment. Turn-finally local pitch events have a higher
likelihood to carry dialog structuring functions: while turn-initial
pitch events are mostly pitch accents, turn-final events often refer
to boundary tones indicating amongst others utterance finality or
continuation. Thus in spoken dialogs they serve as turn-taking and
backchanneling-inviting cues. For both entrainment has been reported
in previous studies by \cite{Gravano2015} and \cite{Levitan2011b},
respectively. Further evidence for entrainment in discourse markers
has been found by \cite{Benus2014}. Thus, one possible explanation for
the higher amount of turn-final as opposed to turn-initial entrainment
is the voluntary dialog structuring influence which adds on to
automatic entrainment especially at the end of turns.

\subsection{Speaker type}

Figures \ref{fig:prob_spk} shows, that describers {\em d\_*} entrain
more than followers {\em f\_*} and among describers, it's the female
speakers {\em d\_f} who entrain. For females {\em *\_f}, describers
entrain more than followers, for males {\em *\_m} it's the opposite.

Globally, disentrainment is to a higher extent found among male
speakers {\em x\_m}, above all among the male describers {\em d\_m}.

Given these findings one can again conclude that entrainment cannot
exhaustively be explained biologically emerging from the perception
behavior-link \cite{Chartrand1999}, since this explanation does not
account for the role-related variation of female and male speakers.

Neither can one conclude that entrainment is a straightforward
function of dominance as predicted by the CAT \cite{Giles2007} in that
sense that the less dominant interlocutor entrains more. Female and
male speakers behave differently in their roles of describers and
followers, describers being equipped with higher authority than
followers due to their lead in knowledge. While men behave in line
with the CAT predictions, i.e. highly disentrain in a high authority
position, female speakers do the opposite.

One motivation for the female behavior might emerge from the
cooperative setting of the game. In this context females might rather
use entrainment to increase communication efficiency instead of
marking authority. However, as shown in Figure \ref{fig:score} for
almost none of the success measures a significant difference between
male and female describers has been observed. Only for the {\em score}
variable we found a significant advantage for male describers.  Thus,
even if the female strategy was to increase communication efficiency,
it was not necessarily successful.

Given the cooperative setting of the game, and the finding that male
speakers did not perform worse in solving the task than female
speakers, it can be concluded that entrainment is used differently
across gender in cooperative solution-oriented interactions. Male
speakers in the role of describers tend to mark hierarchy by
disentrainment, which can be as, or even more, beneficial for task
success as the female strategy of common ground creation by
entrainment.  The amount of entrainment for female and male followers
is about the same. Thus male followers entrain more maybe to signal
that they accept the describer's authority, and female followers
entrain less, since it is less their but rather the describer's
responsibility to establish a common ground.

\section{Conclusion}
\label{sec:conc}
In this paper we set to provide a novel approach to analyzing speech
entrainment in collaborative dialogs. We focused on disentangling the
role of gender and communicative role of the speakers by directed turn
pairing and used novel features for characterizing prosody, an
extended set of analysis units (turn-initial and final IPUs), and a
modified formalization of global and local proximity and synchrony
between interlocutors. The results showed that speech entrainment is a
highly multi-faceted phenomenon as different groups of features show
different entrainment and disentrainment behavior in the local/global
and synchrony/proximity domains. Furthermore, entrainment
predominantly occurs in the turn-final position, which supports a
hybrid account assuming both automatic and voluntary triggers for
entrainment. Finally, the observed gender-role interactions might be
linked to different strategies in solution-oriented collaborative
interactions for males and females.

\section{Acknowledgments}

The work of the first author is financed by a grant of the Alexander
von Humboldt Foundation. This material is based upon work supported by
the Air Force Office of Scientific Research, Air Force Material
Command, USAF under Award no. FA9550-15-1-0055 to the second
author. This work was also funded by the Slovak Scientific Grant
Agency VEGA, grant no. 2/0161/18.

\section*{References}
\bibliographystyle{gerabbrv}
\bibliography{RBM}

\end{document}